\documentclass{article}
\usepackage{spconf,amsmath,graphicx}
\usepackage{booktabs}
\usepackage{multirow}
\usepackage{array}
\usepackage{xcolor}
\usepackage{algorithm}
\usepackage{algorithmic}
\usepackage{url}

\title{Improving Speech-based Emotion Recognition with Contextual Utterance Analysis and LLMs}
\name{Enshi Zhang, Christian Poellabauer}
\address{Florida International University \\
\{ezhan004, cpoellab\}@fiu.edu}

\begin{document}
%
\maketitle
\begin{abstract}
Speech Emotion Recognition (SER) focuses on identifying emotional states from spoken language. The 2024 IEEE SLT-GenSEC Challenge on Post Automatic Speech Recognition (ASR) Emotion Recognition\footnote{\url{https://github.com/YuanGongND/llm_speech_emotion_challenge}} tasks participants to explore the capabilities of large language models (LLMs) for emotion recognition using only text data. We propose a novel approach that first refines all available transcriptions to ensure data reliability. We then segment each complete conversation into smaller dialogues and use these dialogues as context to predict the emotion of the target utterance within the dialogue. Finally, we investigated different context lengths and prompting techniques to improve prediction accuracy. Our best submission exceeded the baseline by 20\% in unweighted accuracy, achieving the best performance in the challenge. All our experiments' codes, prediction results, and log files are publicly available.\footnote{\url{https://github.com/coolsoda/SLT2024_SER}}



\end{abstract}
\section{Introduction}
\label{sec:intro}
One of the key tasks in natural language processing (NLP) and affective computing is Speech-based Emotion Recognition (SER), which focuses on identifying emotional cues in spoken language~\cite{santoso2024large}. Developing advanced algorithms to detect emotions in speech has numerous applications, particularly in the healthcare field. For example, previous work has shown that patients with mental health conditions often exhibit specific emotional patterns in their speech~\cite{tao2024speech,gong2017topic,teodorescu2023language}, which provides opportunities for speech-based psychological assessments and diagnosis.

Creating an effective SER system requires a large amount of labeled emotional speech data. While there is plenty of speech data available, much of it lacks labels. The typical method to label this data is through human annotation, but this process has drawbacks. Emotions are subjective and can lead to biased labels. Even when using domain experts or consensus among multiple annotators to improve reliability, the process is still time-consuming and costly~\cite{latif2023can,gong2023listen,zhang2024mersa}. Large language models (LLMs) offer another solution by automatically annotating emotional speech. Trained on extensive text datasets, LLMs are well-suited for tasks involving ASR~\cite{yu2016automatic} and emotion recognition.

In this paper, we adhere to the requirements and motivations outlined in the GenSEC Task 3~\cite{huck2024large} to investigate the performance of large language models (LLMs) in emotion recognition using only the post-ASR text from the IEMOCAP dataset~\cite{busso2008iemocap}. By analyzing the curated datasets, we aim to showcase the advantages of our approach for this challenge based on the following attributes:

\begin{itemize}
    \item \textbf{ASR refinement:} Due to significant variations in output transcriptions from different ASR models, it is necessary to combine, filter, and refine these outputs to obtain the most informative transcription possible as input data.

    \item \textbf{Dialogues as context:} Unlike the baseline approach, we observed that each session (conversation) consists of smaller dialogue fragments (scripts). This characteristic is present in both the training and testing datasets. Therefore, using these smaller scripts as context to predict the emotion of the target utterance could be more efficient due to relevance and coherence.

    \item \textbf{Prompt structure:} LLMs can consider the entire conversation as context, unlike conventional methods that often treat each sentence separately. While increasing context length could enhance performance, it is essential to determine the optimal length to avoid resource waste. Additionally, the structure of the prompts given to the model — comprising background information, context, current sentences, and tasks — plays a significant role. Exploring prompt engineering techniques can help identify the most effective approach for enhancing prediction accuracy.
\end{itemize}

\section{Related Work}
\begin{figure*}[htbp]
  \centering
  \includegraphics[width=16.0cm]{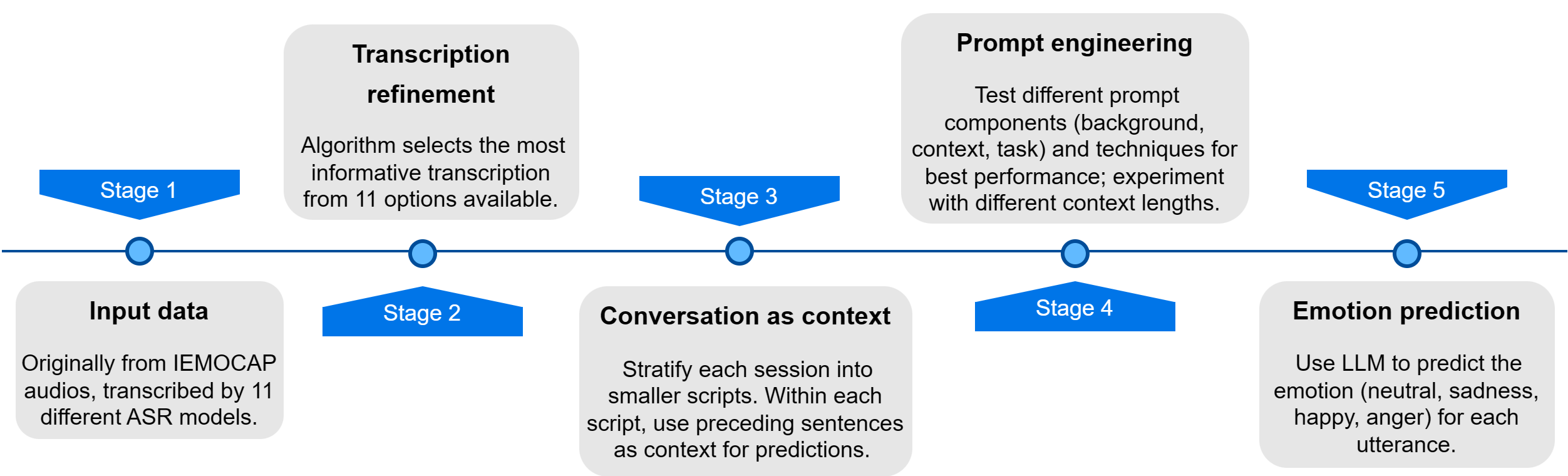}
  \caption{The workflow of our approach.}
  \label{workflow}
\end{figure*}

Extensive prior research has explored the capabilities of LLMs for emotion recognition, often using a multimodal approach. Starting with text data for annotation and downstream SER tasks, many studies found that integrating acoustic features improves performance. For instance, the work in~\cite{li2024speech} benchmarked SER performance using ASR transcripts with varying Word Error Rates (WERs) and explored fusion techniques. They discovered that SER can tolerate relatively low WER, especially in real-world scenarios, and that bimodal SER with transcripts containing about 10\% errors can perform comparably to those with ground-truth text, particularly with Whisper models. They also proposed an ASR error-robust framework integrating ASR error correction and modality-gated fusion.

The work in~\cite{feng2024foundation} investigated using foundational models to automate SER from transcription and annotation. They used these models for emotion annotation, combining outputs from multiple LLMs and incorporating human feedback to reduce bias. Their experiments showed that using both speech and text led to higher accuracy compared to using speech alone. Additionally, LLM-generated transcriptions, such as those from Whisper-large V2, were competitive with human-transcribed ground truth. The highest prediction accuracy was achieved by combining LLM outputs and incorporating human feedback to enhance zero-shot capabilities. Similarly, the work in~\cite{santoso2024large} demonstrated that LLMs could effectively annotate emotional speech through prompted transcriptions. Their SER experiments using these annotations achieved competitive results compared to ground truth labels. They also explored providing both conversation text and acoustic features to LLMs for annotations. In~\cite{latif2023can}, the authors examined the potential of LLMs to annotate large volumes of speech data. By evaluating single-shot and few-shot scenarios using the IEMOCAP, MSP-IMPROV, and MELD datasets, they observed performance variability in SER. They developed a technique using ChatGPT to annotate publicly available data for training data augmentation, which improved performance on downstream SER tasks.

There are also studies that focus exclusively on text-based methods. The work in~\cite{gong2023lanser} investigated LLMs for annotating emotional speech data to reduce the human burden. They extracted text transcripts using ASR, then fed the transcriptions into pre-trained LLMs to infer weak emotion labels using an engineered prompt and a predetermined, finer-grained emotion taxonomy. These labels were used to pre-train an SER model via weakly-supervised learning, significantly improving SER performance on unlabeled data and enhancing label efficiency. In ~\cite{amin2024prompt}, the authors evaluated the prompting and sensitivity of the foundation model ChatGPT to determine how different prompting templates and sampling sensitivity affect the correctness and helpfulness of the responses. They conducted three affective computing tasks: sentiment analysis, toxicity detection, and sarcasm detection. By evaluating various prompting techniques, they demonstrated that straightforward prompts and assigning expert identity yield near-best performances. They also highlighted that irrelevant expertise or misaligned incentives can severely harm results.


\section{Dataset}
For this emotion recognition task, we use the IEMOCAP dataset. It contains dialogues from 10 actors (5 male and 5 female) divided into five extensive sessions to convey target emotions. The complete dataset includes audio of both scripted and spontaneous dialogue scenarios, visual data capturing facial expressions and hand movements, and transcribed text. The emotions are labeled discretely as happy, anger, sadness, frustration, and neutral, as well as categorically on scales of valence, arousal, and dominance. Human transcribers and annotators have labeled and validated these as ground truth. For this challenge, we use only the post-ASR texts to predict emotions for each utterance, with the target output emotion being neutral, sad, happy, or angry.

\section{Methodology}
Before starting our analysis, we thoroughly reviewed the challenge rules and closely examined both the training and evaluation data provided. The organizers had already curated the training and testing data in JSON format for direct use. We identified several key findings that informed our proposed methods. We discuss each of these findings in detail below (also see Figure~\ref{workflow}).

\subsection{Transcription refinement}
\label{refinement approach}
\begin{table}[t]
    \centering
    \large
    \resizebox{\columnwidth}{!}{
    \begin{tabular}{ll}
        \toprule
        \textbf{Key} & \textbf{Value}\\
        \midrule
        need\_prediction & yes \\
        emotion & sad \\
        id & Ses01F\_script01\_3\_M023 \\
        speaker & Ses01\_M \\
        \textbf{Ground truth} & `Yeah. I suppose I have been. But it's going from me.' \\
        hubertlarge & `ya i suppose i have been bht's going from me' \\
        w2v2100 & `a i suppose i have been but's going from me' \\
        w2v2960 & `oh i suppose i have been let's going from me' \\
        w2v2960large & `now i suppose i have been bat's going from me' \\
        w2v2960largeself & `ar i suppose i have been but's going from me' \\
        wavlmplus & `a i spose a habben was going for m' \\
        whisperbase & `Yeah' \\
        whisperlarge & `Yeah' \\
        whispermedium & `Yeah' \\
        whispersmall & `Yeah' \\
        whispertiny & `Yeah' \\
    
        \bottomrule
    \end{tabular}}
    \caption{A sample data entry from the training set, displaying all key-value pairs.}
    \label{trainingsample}
\end{table}
By examining several data entries in the training and testing datasets, we observed significant variation in the transcriptions provided by different ASR models. Table~\ref{trainingsample} shows a sample data entry from the training set, which includes the ground truth transcription. As illustrated, some models provide extremely limited information. Inspired by previous work~\cite{li2022fusing}, we conducted a Word Error Rate (WER) analysis on the training dataset for this challenge (Table~\ref{worderrorrate}), revealing inconsistencies across the models. To address this issue, we designed a straightforward algorithm (Algorithm~\ref{alg:transcription_refinement}), that, for each data entry, filtered out any transcriptions that are too short to provide enough information. If all transcriptions are short, we kept them all. Next, we refined the remaining transcriptions using ChatGPT 3.5. This refinement process selects the most coherent and logically structured transcription. If no clear choice is possible, we select the longest transcription, as it likely contains the most information. This process generates a new key for each data entry, which we call `\textbf{ensemble}.' We expect this approach to improve data quality for subsequent predictions.

\subsection{Script-based}
\label{script approach}
\begin{table}[t]
    \centering
    \resizebox{\columnwidth}{!}{
    \begin{tabular}{lcccccc}
        \toprule
        & \textbf{Neutral} & \textbf{Sad} & \textbf{Happy} & \textbf{Angry} & Other & \textbf{Overall}\\
        \midrule
        hubertlarge & 0.20 & 0.18 & 0.25 & 0.18 & 0.19& 0.20 \\
        w2v2100 & 0.41 & 0.34 & 0.43 & 0.37 & 0.36& 0.38 \\
        w2v2960 & 0.35 & 0.28 & 0.33 & 0.26 & 0.28&  0.29 \\
        w2v2960large & 0.31 & 0.24 & 0.29 & 0.22 & 0.24& 0.25 \\
        w2v2960largeself & 0.23 & 0.18 & 0.22 & 0.15 & 0.18& 0.19 \\
        wavlmplus & 0.45 & 0.39 & 0.43 & 0.32 & 0.37& 0.38 \\
        whisperbase & 0.36 & 0.40 & 0.48 & 0.34 & 0.38& 0.39 \\
        whisperlarge & 0.36 & 0.41 & 0.46 & 0.35 & 0.36& 0.38 \\
        whispermedium & 0.33 & 0.37 & 0.45 & 0.34 & 0.35& 0.36\\
        whispersmall & 0.35 & 0.39 & 0.47 & 0.33 & 0.36& 0.37 \\
        whispertiny & 0.50 & 0.42 & 0.55 & 0.42 & 0.41& 0.44 \\
        \midrule
        Utterance & 607 &475 & 681 & 814 & 2648 & 5225 \\
    
        \bottomrule
    \end{tabular}}
    \caption{Word Error Rate (WER) analysis for all data entries with emotion labels in the training set. \textit{Utterance} represents the number of instances for each emotion label.}
    \label{worderrorrate}
\end{table}

Both the training and testing data are based on five main sessions (conversations). The baseline proposed by the organizers suggests using the entire conversation as context, with a default context length of 3, to predict the emotion of the target utterance. However, we observed that each conversation consists of smaller dialogues (scripts) which can be processed individually. Thus, we believe it is more efficient to use these scripts as context rather than the entire conversation.

\begin{algorithm}[H]
\caption{Pseudo-code for ASR transcriptions refinement}
\label{alg:transcription_refinement}
\begin{algorithmic}
\STATE \textbf{Input:} List of transcriptions from various ASR models
\STATE \textbf{Output:} Refined transcription

\STATE Define \textit{min\_length} = 5

\STATE

\STATE \textbf{1. Function} GetFilteredTranscriptions \newline
(transcriptions, \textit{min\_length})
\STATE \quad filtered\_transcriptions $\gets$ only keep transcriptions longer than \textit{min\_length}
\STATE \quad \textbf{if} filtered\_transcriptions is empty \textbf{then}
\STATE \quad \quad \textbf{return} all\_transcriptions
\STATE \quad \textbf{else}
\STATE \quad \quad \textbf{return} filtered\_transcriptions
\STATE \quad \textbf{end if}

\STATE

\STATE \textbf{2. Function} RefineTranscriptionWithGPT \newline
(transcriptions\_filtered)
\STATE \quad Construct prompt for GPT-3.5 as: ``You are a text refinement assistant. Choose the most comprehensive and coherent sentence from the following options. If impossible to decide, choose the longest option available. Output only the selected sentence without any additional explanation or phrases."
\STATE \quad response $\gets$ call GPT-3.5 with prompt
\STATE \quad \textbf{return} response

\end{algorithmic}
\end{algorithm}

\begin{table*}[t]
\small
    \centering
    \renewcommand{\arraystretch}{1.3}
    \begin{tabular}{p{2.1cm}|p{13.8cm}}
    \toprule
         \textbf{Name}& \textbf{Prompt Template}\\
         
         \midrule
         Baseline& Two speakers are talking. The conversation is \{context\}. Now speaker \{current speaker\} says: \{current sentence\}. Predict the emotion of the sentence \{current sentence\} from the options [happy, sad, neutral, angry], consider the conversation context, do not explain, only output the label in [happy, sad, neutral, angry].\\

         Expert& You are an expert emotion predictor. Analyze the emotions based on the conversation context \{context\} provided. Now speaker \{current speaker\} says: \{current sentence\}. Predict the emotion of the sentence \{current sentence\} from the options [happy, sad, neutral, angry], consider the conversation context, do not explain, only output the label in [happy, sad, neutral, angry].\\

         Gambler& You are an expert gambler who earns money by predicting emotions correctly. Your goal is to maximize profit by making the best possible predictions. Two speakers are talking. The conversation is \{context\}. Now speaker \{current speaker\} says: \{current sentence\}. Predict the emotion of the sentence \{current sentence\} from the options [happy, sad, neutral, angry], consider the conversation context, do not explain, only output the label in [happy, sad, neutral, angry].\\

         CoT& You are an expert emotion predictor. Analyze the emotions based on the conversation context \{context\} provided. Now speaker \{current speaker\} says: \{current sentence\}. Based on the context provided and the current sentence, follow these steps: 1. Analyze the conversation context and note any emotional cues. 2. Consider how these cues might affect the current speaker's emotion. 3. Predict the emotion of the current sentence from the options [happy, sad, neutral, angry]. 4. Do not explain your reasoning, only output the label in [happy, sad, neutral, angry]. \\

         CoT-fired& You are an expert emotion predictor. Analyze the emotions based on the conversation context \{context\} provided. Now speaker \{current speaker\} says: \{current sentence\}. Based on the context provided and the current sentence, follow these steps: 1. Analyze the conversation context and note any emotional cues. 2. Consider how these cues might affect the current speaker's emotion. 3. Predict the emotion of the current sentence from the options [happy, sad, neutral, angry]. 4. Do not explain your reasoning, only output the label in [happy, sad, neutral, angry]. If you do not get the prediction right, I will be fired and lose my job. So please try you best.\\
    
    \bottomrule
    \end{tabular}
    \caption{The various prompts tested on the training dataset. Each prompt includes background information, context information (previous utterances up to the defined context length), the current sentence (current speaker and current sentence), and the task (what to predict and the rules to follow).}
    \label{Prompt template}
\end{table*}

For example, assume that Session 01, Script 04 has 20 utterances (01-04-01 to 01-04-20) and the next utterance to be predicted is Session 01, Script 05, Utterance 03 (01-05-03). Following the baseline approach with a context length of 3, we would consider utterances 01-05-02, 01-05-01, and 01-04-20 as context. However, 01-04-20 belongs to the previous script, which could contain irrelevant context and potentially harmful to the prediction's accuracy. Therefore, we limit the context to previous utterances within the same script. In this case, we would only use 01-05-02 and 01-05-01 as context, even with context length of 3. It ensures that all utterances within the same script will not interfere with predictions in the next script.

\subsection{Prompting}
\label{prompting techniques}
Following the baseline method, we utilize ChatGPT through the official API\footnote{\url{https://platform.openai.com/docs/api-reference/introduction}}. ChatGPT's sensitivity to prompt variability can lead to ambiguous and erroneous results with slight changes to the prompt content. This is particularly challenging because human emotions in short dialogues, like those in our training and evaluation datasets, often exhibit significant variability~\cite{latif2023can}. To determine the most effective prompting techniques for this task, we conducted extensive experiments inspired by previous research~\cite{latif2023can,amin2024prompt,santoso2024large}. We explored the impact of mentioning subject-matter expertise in the prompts and tested the effects of mentioning irrelevant expertise or incentives to provide correct responses. Additionally, we included additional details to structure the answer and utilized a step-by-step approach known as Chain-of-Thought (CoT). Another approach combined extra details, step-by-step thinking, and mentioned penalties for incorrect answers.

The prompt templates are in Table~\ref{Prompt template}. Each prompt includes background information, context (previous sentences up to the defined context length), the current sentence (current speaker and current sentence), and the task (predictions and requirements). This suite of prompts allows for a comprehensive evaluation of how prompt design influences the performance of LLMs.

\section{Experiments and Results}
\subsection{Strategy}
\begin{table*}[t]
\centering
\renewcommand{\arraystretch}{1.2} 
\begin{tabular}{l|cccc|c|c}
    \hline
    \multirow{2}{*}{Methods} & \multicolumn{4}{c|}{F1 score (train)} & \multirow{2}{*}{UA (train)} & \multirow{2}{*}{UA (test)} \\
    \cline{2-5}
     & Neutral & Sadness & Happy & Anger &  \\
    \hline
    \textbf{1. whispertiny + baseline\_3 + session-based (baseline)}& 0.44 & 0.37 & 0.23 & 0.61 & \textbf{0.446} & \textbf{0.551} \\
    
    \textcolor{blue}{2. w2v2960largeself + baseline\_3 + session-based} & 0.43 & 0.50 & 0.40 & 0.66 & 0.511 & \textcolor{blue}{0.575} \\
    
    3. w2v2960largeself + baseline\_3 + script-based & 0.43 & 0.50 & 0.40 & 0.67 & 0.514 &-- \\
    
    \midrule
    4. ensemble + baseline\_3 + session-based & 0.43 & 0.49 & 0.36 & 0.67 & 0.504 &-- \\
    5. ensemble + baseline\_3 + script-based & 0.43 & 0.48 & 0.40 & 0.68 & 0.510 &--\\
    \midrule
    \textcolor{blue}{6. ensemble + baseline\_5 + script-based} & 0.44 & 0.52 & 0.40 & 0.70 & 0.529 & \textcolor{blue}{0.659} \\
    \textcolor{blue}{7. ensemble + baseline\_10 + script-based} & 0.45 & 0.55 & 0.44 & 0.70 & 0.548 & \textcolor{blue}{0.694} \\
    8. ensemble + baseline\_15 + script-based & 0.47 & 0.57 & 0.48 & 0.70 & 0.562 &-- \\
    9. ensemble + expert\_10 + script-based & 0.44 & 0.50 & 0.46 & 0.69 & 0.537 &-- \\
    10. ensemble + gambler\_10 + script-based & 0.38 & 0.54 & 0.48 & 0.69 & 0.542 &-- \\
    11. ensemble + CoT\_10 + script-based & 0.44 & 0.50 & 0.39 & 0.66 & 0.510 &-- \\
    12. ensemble + CoT-fired\_10 + script-based & 0.45 & 0.52 & 0.34 & 0.62 & 0.491 &-- \\
    \midrule
    \textcolor{blue}{\textbf{13. ensemble + baseline\_10 + script-based}} & 0.50 & 0.59 & 0.32 & 0.73 & 0.559 &\textcolor{blue}{\textbf{0.752}} \\
    \hline
\end{tabular}
\caption{Summary of all methods tested. Methods highlighted in blue were applied to the test set. The method highlighted in blue and bold is our best performing one, utilizing GPT-4.0, while all other methods, including the baseline, are based on GPT-3.5-turbo.}
\label{Results: training}
\end{table*}

In this challenge, only the training data contains accurate transcriptions and emotion labels. The performance on the testing dataset will only be available after submitting the predicted labels. Given the limited number of allowed submissions, we first conducted all our experiments on the training set. Initially, here is our plan: after the first submission, we will assess whether the accuracy improves by switching from `whispertiny' to `w2v2960largeself'. If so, we are moving in the right direction by improving WER on input data. Then, we can try other methods that have higher unweighted accuracy (UA) than the approach used in the first submission.

However, a few considerations came to mind. First, the WER may differ between the train and test set. From the training set, we have observed significant variations in transcriptions from ASR models across data entries. This may also hold true for the test set. Therefore, we choose to use the methods that takes `ensemble' as input for subsequent submissions, even though it lowered the UA between experiments 2 and 4 and between 3 and 5. Second, each data entry in the train and test set slightly differs in how each conversation was divided into smaller `scripts.' For instance, in the training set, the data entries with id `Ses01F\_script01\_1\_F000' and `Ses01F\_script01\_2\_F000' are both considered under script01, even though they belong to different subsets of the same script. In the test set, the data entry have ID such as `Ses01Z\_01\_F000', which clearly belong to script 01, so `Ses01Z\_02\_F000' belongs to the second script. There are no further subsets. This difference could impact our prediction because our approach processes each script individually rather than the more nuanced subsets of the script. This may also explain why the prediction only slightly improved when we switched from a session-based to a script-based approach between experiments 2 and 3 and 4 and 5. Based on these considerations, we use the method that takes `ensemble' as input for subsequent submissions, even though it lowered the UA between experiments 2 and 4 and between 3 and 5. Based on these considerations, we use the 'ensemble' transcription as input and a script-based approach for subsequent experiments.

\subsection{Metrics}
For evaluation, we used unweighted accuracy (UA), as required by the challenge. Additionally, since we have the ground truth for the training data, we employed the F1 score to assess the model's performance in predicting different emotional labels.

\subsection{Results}
Table \ref{Results: training} presents all the methods we tested, including the one used to reproduce the baseline method. In total, we conducted 13 experiments. Each method consists of three components: the transcription used, the prompt and context length, and whether the entire session or a shorter script was used as context. In our first submission, the only difference from the given baseline was the transcription used. We selected the transcription from `w2v2960largeself', which had the lowest WER among all options. This improved all metrics on the training set and already outperformed the baseline on the testing set. Recognizing the importance of accurate transcription, we adopted the transcription refinement approach (Section~\ref{refinement approach}), which further improved our results.

Next, we applied the script-based approach (Section~\ref{script approach}), which slightly enhanced performance. Combining this with the ensemble approach, we confirmed that the script method outperformed the session-based approach, so we adhered to the script approach.

We then focused on prompting techniques. Initially, we found that increasing context length improved performance. Keeping other prompt components the same, we tested context lengths of 3 (baseline), 5, 10, and 15. There was a steady increase in accuracy until 10, with no significant improvement from 10 to 15. Thus, we settled on using a context length of 10. Following the prompting techniques discussed in Section~\ref{prompting techniques}, we conducted four more experiments. However, they all performed worse compared to the baseline prompt.

\section{Discussion}
In our experiments, we found that the best results were obtained by using refined transcription with a baseline prompt of context length 10 and using the script as context to predict the emotion of the target utterance. Based on the WER analysis, we chose the ASR output by `w2v960largeself', which had the lowest overall WER of 0.19, compared to the baseline `whispertiny' with an overall WER of 0.44. The unweighted accuracy increased from 0.446 to 0.511 on the training set. The results improved further when we used an `ensemble' approach to filter, combine, and refine all available transcriptions, emphasizing the importance of WER in the task of SER.

Switching from a session-based approach to a script-based approach has resulted in a slight improvement in our prediction accuracy between experiments 2 and 3, as well as between experiments 4 and 5. We anticipate that this change will have a greater impact on the testing set since it does not have subsets of scripts like the training set data.

The role of context length in prediction performance is vital. Simply increasing the context length from 3 to 5 and then from 5 to 10 significantly improved accuracy. This finding helps us understand context length's impact on prediction accuracy. It also highlighted LLMs' ability to use longer context for making predictions, compared to the conventional method that treats each sentence in a sequence of conversations separately. However, it is essential to notice that performance does not improve when we increase context length from 10 to 15. The reasons could be due to the data structure (short dialogues in general). For instance, this could change when working on monologues or reading-based datasets. It is essential to be aware of the limitations of LLMs when using extended context to make predictions.

By assigning different roles, incentives, and penalty information, we explored different prompting techniques to see if LLMs are more sensitive to a specific prompt. However, the simple baseline prompt performed better. This result differed from previous approaches~\cite{amin2024prompt} investigating different prompt effectiveness on sentiment analysis.

From the F1 scores on training sets, we found that utterances with the emotion label `happy' are the hardest to predict compared to others. Anger is the easiest to predict. Models have difficulty distinguishing between neutral and sad and between neutral and happy. These emotional shifts during a conversation are still challenging for LLMs to detect through text alone.

In our latest experiment, we utilized the GPT 4.0 for the first time due to limited resources available. By applying our best method on the test set, the unweighted accuracy improved from 0.694 to 0.752. From the feedback after submission, we observed an improvement in predictions for all emotion labels except for the `happy' emotion. In comparison to the GPT 3.5 turbo model, the GPT 4.0 model tends to classify more `happy' instances as `neutral.' Despite this, the overall performance of the GPT 4.0 model is much better.

\section{Conclusions}
SER is an essential area of study that has significantly contributed to various multimodal analyses. In this challenge, participants were required to use post-ASR transcriptions and LLMs to predict the emotion conveyed in each utterance, categorizing each as neutral, sad, happy, or angry. The dataset was organized by the order of the session (conversation).

In our approach, we applied a refinement algorithm to all ASR transcriptions to select the most informative one as input. We also identified that each conversation could be broken down into smaller dialogues (scripts). Using these scripts as context, we predicted the emotion of the target utterance. We explored using different context lengths and prompting techniques in addition to these two strategies to improve prediction accuracy further. We developed a method that achieved higher unweighted accuracy than the baseline using the ChatGPT 3.5 turbo model. The result was further enhanced when we applied the same method to ChatGPT 4.0.

The results confirmed that LLMs can effectively consider more extended contexts preceding the current sentence to predict emotions, unlike conventional methods that often treat each sentence separately. Additionally, we found that the emotion of anger is the easiest for LLMs to predict accurately while distinguishing other emotions, particularly happy and neutral, can be more challenging.





\bibliographystyle{IEEEbib}
\bibliography{myreferences}

\end{document}